\documentclass{article}
\usepackage{spconf,amsmath,graphicx,amssymb}
\usepackage{booktabs}
\usepackage{color}

\title{Complementary Network with Adaptive Receptive Fields\\ for Melanoma Segmentation}
\name{Xiaoqing Guo, Zhen Chen, Yixuan Yuan \thanks{This work was supported by CityU TSG 6000690. We gratefully acknowledge the support of NVIDIA Corporation with the donation of the Titan Xp GPU for this research.}}
\address{Department of Electrical Engineering, City Univeristy of Hong Kong, Hong Kong SAR, China}

\begin{document}
%
\maketitle
\begin{abstract}
Automatic melanoma segmentation in dermoscopic images is essential in computer-aided diagnosis of skin cancer. Existing methods may suffer from the hole and shrink problems with limited segmentation performance. To tackle these issues, we propose a novel complementary network with adaptive receptive filed learning. Instead of regarding the segmentation task independently, we introduce a foreground network to detect melanoma lesions and a background network to mask non-melanoma regions. 
Moreover, we propose adaptive atrous convolution (AAC) and knowledge aggregation module (KAM) to fill holes and alleviate the shrink problems.
AAC explicitly controls the receptive field at multiple scales and KAM convolves shallow feature maps by dilated convolutions with adaptive receptive fields, which are adjusted according to deep feature maps. 
In addition, a novel mutual loss is proposed to utilize the dependency between the foreground and background networks, thereby enabling the reciprocally influence within these two networks.
Consequently, this mutual training strategy enables the semi-supervised learning and improve the boundary-sensitivity.
Training with Skin Imaging Collaboration (ISIC) 2018 skin lesion segmentation dataset, our method achieves a dice coefficient of 86.4\% and shows better performance compared with state-of-the-art melanoma segmentation methods\footnote{{\url{https://github.com/Guo-Xiaoqing/Skin-Seg}}}.

\end{abstract}
\vspace{-0.2cm}
\section{Introduction}
\vspace{-0.1cm}
Melanoma is the most dangerous form of skin cancer, accounting for a large percentage of skin cancer deaths \cite{siegel2019cancer}. 
Fortunately, if detected early, melanoma survival rate exceeds 95\% \cite{siegel2019cancer}.
Dermoscopy is an imaging technique to visualize deep levels of the skin and it is widely applied to diagnose melanoma.
However, manually reviewing dermoscopy images is an error-prone and time-consuming work even for professional dermatologists. 
In this regard, the development of computational support systems for automated segmentation and analysis of dermoscopy images is highly desirable.

\begin{figure}[!tp]
\centering
\includegraphics[width=85mm]{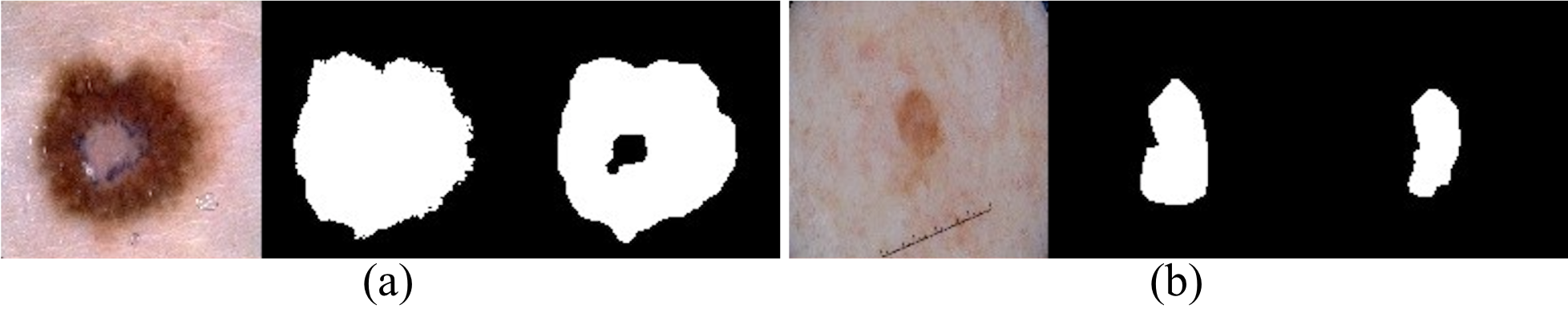}
\vspace{-0.2cm}
\caption{Illustrations of (a) hole problem, (b) shrink problem. Each group includes the original image, ground truth and prediction of U-Net \cite{ronneberger2015u} from left to right.} 
\label{problem}
\vspace{-0.6cm}
\end{figure}

Automated melanoma segmentation remains to be challenging 
due to the huge variations of melanomas in terms of shape, color and texture. Moreover, some samples may contain artifacts, such as hairs, ruler marks and color calibrations, blurring melanoma lesions. Many algorithms have been proposed to tackle these challenges \cite{li2019dense,sarker2018slsdeep,yuan2017improving}.
Yuan et al. \cite{yuan2017improving} incorporated Lightness channel from CIELAB color space and three channels of HSV space together for the melanoma segmentation. 
Sarker et al. \cite{sarker2018slsdeep} presented a melanoma segmentation model with negative log likelihood and end point error loss functions to preserve sharp boundaries. 
Li et al. \cite{li2019dense} proposed a dense deconvolutional network with hierarchical supervision to capture local and global contextual information for melanoma segmentation. 
Although existing methods have achieved significant success, they still suffer from the hole (Fig.\ref{problem} (a)) and shrink (Fig.\ref{problem} (b)) in predictions. The relatively low contrast between melanoma and non-melanoma regions confuses the network and causes the appearance of holes. The fuzzy boundaries lead to the shrinking prediction and further decrease the sensitivity of prediction. 

\begin{figure*}[!tp]
\centering
\includegraphics[width=175mm]{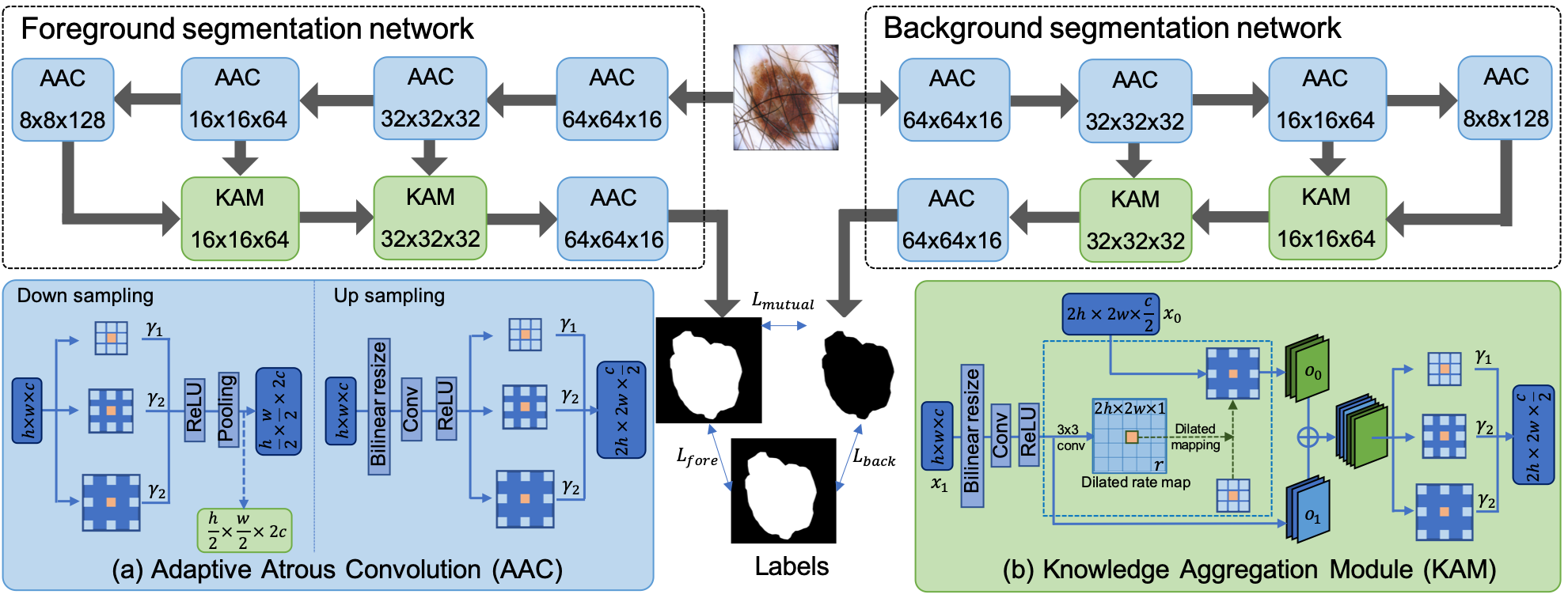}
\caption{Illustration of the proposed complementary segmentation network, including foreground and background networks.}
\vspace{-0.2cm}
\label{framework}
\vspace{-0.2cm}
\end{figure*}

To address the hole and shrink problems mentioned before, we propose a complementary network consisting of a foreground segmentation network and a background segmentation network. With the fact that the dependency of two networks is crucial, we propose a mutual loss to optimize the complementary network collaboratively. In this way, our network is sensitive to boundary and can effectively deal with shrink problem. 
Additionally, we propose AAC to explicitly control the receptive field for incorporating local and context information, and KAM to convolve shallow features by adaptive receptive field kernels learned from deep features. With AAC and KAM, our model can expand the segmented region to fit the ground truth lesion and fill holes.

\vspace{-0.2cm}
\section{Method}
\vspace{-0.1cm}
In this paper, we propose a complementary network for melanoma segmentation in Fig.\ref{framework}. Our model is composed of two networks, foreground and background segmentation networks. 
For each segmentation network, the input image is firstly passed through the contracting path with four downsampling AACs.
The sequences of AACs extract multi-scale features from low-level contexture details to high-level semantic structures. Then, extracted features from different stages are incorporated by KAMs, which can aggregate informative features and suppress noises. After that, an upsampling AAC is introduced to further extend the size of feature maps. Finally, an upsampling layer and a fully connected layer are used to predict the class result per pixel. The foreground segmentation network and background network are trained collaboratively and jointly by minimizing the proposed mutual loss together with their individual foreground and background loss. 

\vspace{-0.2cm}
\subsection{Adaptive Atrous Convolution}
Atrous convolution \cite{li2019dense, sarker2018slsdeep} allows us to explicitly enlarge the receptive field of filters. 
However, it is tricky to manually choose the dilated rate, and the fixed-size receptive field is the main limitation of atrous convolution. We propose the AAC module to alleviate information loss caused by sparse kernels and 
adaptively enlarge the receptive field as in Fig.\ref{framework} (a). We first utilize various dilated rates to enlarge the receptive field of kernels. 
Considering different sizes and shapes of objects in images, the importance of feature maps with different receptive fields may not be equal.
Therefore, we assign an importance score for each output of dilated convolution, and AAC can be formulated as:

\vspace{-0.2cm}
\begin{footnotesize}
\begin{equation}
g[i, j] = \sum_{k=1}^{K}\sum_{m=1}^{M}\sum_{n=1}^{N}\gamma_{k}\cdot f[i+r_{k}\cdot m, j+r_{k}\cdot n]\cdot h[m, n],
\end{equation}
\end{footnotesize}where $f[i, j]$ and $g[i, j]$ are input and output feature maps of AAC. $h[m, n]$ denotes the convolution kernel with $M\times N$ size. $r_{k}$ is dilated rate that equals to $k$, and $K$ is set as 3. $\gamma_{k}$ denotes the important score for dilated convolution with rate $r_{k}$. All $\gamma$ values are initialized as $\frac{1}{3}$ and updated every iteration during training phases. Thus, each layer can adjust to its appropriate receptive fields gradually. Finally, the weighted output feature maps are added together for further processing. With receptive fields being adaptively enlarged, the context information can be incorporated into local information, and the problem of holes and shrinks can be relieved. 

\vspace{-0.2cm}
\subsection{Knowledge Aggregation Module}
Previous methods usually use the ``skip-connection" to concatenate multi-level features directly for accurate segmentation \cite{li2019dense, ronneberger2015u, sarker2018slsdeep}. 
However, the equal weights for different channels of concatenated features are deficient due to the redundant information. Moreover, it is necessary to incorporate context with local information to fill holes since the context information may provide rich clues for local prediction. Therefore, we propose KAM to amalgamate and distill multi-level features as in Fig.\ref{framework} (b). 
Assume feature maps from last layer are $x_{1} \in \mathbb{R}^{h\times w\times c}$ and 
features from the contracting path are $x_{0} \in \mathbb{R}^{2h\times 2w\times \frac{c}{2}}$, $x_{1}$ is first convoluted to $o_{1} \in \mathbb{R}^{2h\times 2w\times \frac{c}{2}}$. 
Then, a $3\times 3$ convolution layer, rate learning layer, is applied to $o_{1}$ to learn dilated rate map $r \in \mathbb{R}^{2h\times 2w\times 1}$. Each pixel in $r$ indicates the dilated rate of convolution kernel at the corresponding position, and the dilated rate map controls the scaling of receptive fields for each position individually. The weight of the rate learning layer is initialized as $N(0, \sigma ^{2})$ with $\sigma \ll 1$, and the bias is initialized as ones. This initialization method makes the convolution kernel start from the standard convolutions and gradually adjust to the appropriate dilated rate. Through dilated mapping, the learned dilated rate are applied to each position for convolving $x_{0}$, which can be formulated as:

\vspace{-0.3cm}
\begin{footnotesize}
\begin{equation}
o_{0}[i, j] = \sum_{m=1}^{M}\sum_{n=1}^{N}x_{0}[i+r[i, j]\cdot m, j+r[i, j]\cdot n]\cdot h[m, n].
\end{equation}
\end{footnotesize}When the coordinates $(i+r[i, j]\cdot m, j+r[i, j]\cdot n)$ are not at grids, bilinear interpolation is utilized for approximation as in \cite{jaderberg2015spatial}. Finally, the output $o_{0}$ of adaptive dilated convolution is concatenated with $o_{1}$ for further processing. 
Compared with the standard convolution operator, this dilated mapping enables flexible-size receptive fields according to the semantic information. Moreover, KAM incorporates context information from deep semantic features to texture details, which can alleviate the hole problem and suppress noises.

\vspace{-0.2cm}
\subsection{Joint Loss Function}
\subsubsection{Foreground and Background Loss.} 

We use focal loss \cite{lin2017focal} and Jaccard loss \cite{yuan2017improving} to prevent the problem of foreground-background class imbalance. 
Assume $y_{ij}$ is the ground truth of pixel $j$ in image $i$, its corresponding probability predicted for correct class can be calculated by $p_{ij}^{f} = \frac{e^{W^{\top}_{y_{ij}}x_{ij}+b_{y_{ij}}}}{\sum_{k=1}^{2}e^{W^{\top}_{k}x_{ij}+b_{ij}}}$, where $W$ and $b$ are weights and bias in the fully connected layer. Then loss function of foreground network can be formulated as :

\vspace{-0.2cm}
\begin{footnotesize}
\begin{equation}
\begin{split}
L_{fore} = -\sum_{i=1}^{N}\sum_{j=1}^{n} \frac{(1-p_{ij}^{f})^{2}\log(p_{ij}^{f})}{N \times n} + \frac{p_{ij}^{f}y_{ij}}{p_{ij}^{f} + y_{ij} - p_{ij}^{f}y_{ij}},
\end{split}
\end{equation}
\end{footnotesize}where $N$ denotes mini-batch size, and $n$ denotes the number of pixels in a dermoscopy image. The first term in $L_{fore}$ is focal loss, while the second one is Jaccard loss. 

In background segmentation network, $1-y_{ij}$ is ground truth of background pixel $j$ in image $i$. 
Similar to foreground loss, background loss function is calculated as:

\vspace{-0.2cm}
\begin{footnotesize}
\begin{equation}
\begin{split}
L_{back}=-\sum_{i=1}^{N}\sum_{j=1}^{n} \frac{(1-p_{ij}^{b})^{2}\log(p_{ij}^{b})}{N \times n}+\frac{p_{ij}^{b}(1-y_{ij})}{(1-y_{ij})+p_{ij}^{b}y_{ij}}.
\end{split}
\end{equation}
\end{footnotesize}This loss function can alleviate class imbalance problem, and avoid additional procedures to re-balance pixels from melanoma region and background.

\vspace{-0.2cm}
\subsubsection{Mutual Loss}
To exploit complementary information among the foreground network and background network, we introduce a constraint by minimizing the similarity of predictions from two networks.
In particular, we utilize Jensen-Shannon (JS) divergence to measure the similarity of predictions from two networks and propose an exclusion loss to enforce the predicted segmentations from two networks mutually exclusive. 
The mutual loss thus can be formulated as: 

\vspace{-0.2cm}
\begin{footnotesize}
\begin{equation}
\begin{split}
\mathit{L}_{mutual}=\frac{1}{N\times n}\sum_{i=1}^{N}\sum_{j=1}^{n}\frac{p^{f}_{ij}}{2}\log \frac{2p^{f}_{ij}}{p^{f}_{ij}+(1-p^{b}_{ij})}+\\ \frac{1-p^{b}_{ij}}{2}\log \frac{2(1-p^{b}_{ij})}{p^{f}_{ij}+(1-p^{b}_{ij})}+\frac{2 p_{ij}^{0}p_{ij}^{1}}{p_{ij}^{0}+p_{ij}^{1}},
\end{split}
\end{equation}
\end{footnotesize}where the first and second terms in $L_{mutual}$ are JS loss, and the third one is the exclusion loss. By minimizing the mutual loss, distributions of $p^{f}_{ij}$ and $1-p^{b}_{ij}$ are constrained to be similar, and the overlap of predictions from two networks tends to be minimized. Mutual loss enhances the complementary information within these two networks.  Moreover, the prediction around the boundary is prone to agree with ground truth, and the problem of shrinks will be alleviated. 

\vspace{-0.2cm}
\subsubsection{Extension to Semi-supervised Learning}
The proposed complementary network can be extended to semi-supervised learning. Under the semi-supervised learning setting, we use foreground loss and background loss for labeled data and compute mutual loss for all training data. 
Denote the labeled and unlabeled data as $\mathcal{L}$ and $\mathcal{U}$, the total loss function can be represented as

\begin{footnotesize}
\begin{equation}
L_{Total} = \mathop {L_{fore}}\limits_{x\in \mathcal{L}} + \mathop {L_{back}}\limits_{x\in \mathcal{L}} + \mathop {L_{mutual}}\limits_{x\in \mathcal{D}},
\end{equation}
\end{footnotesize}where $\mathcal{D} = \mathcal{L}\cup\mathcal{U}$. Therefore, the complementary network can be not only optimized with pixel-level annotation, but also supervised by dual networks collaboratively and jointly. 

\vspace{-0.2cm}
\section{Experiments and Results}
We evaluated the proposed method on 2018 ISIC skin lesion segmentation dataset \cite{Codella2018Skin}, which includes 2594 annotated dermoscopic images.
Fourfold cross validation was adopted for the evaluation. The performance of segmentation was evaluated by accuracy (AC), dice coefficient (DI), Jaccard index (JA) and sensitivity (SE). 
We implemented our model with TensorFlow, and NVIDIA TITAN XP GPU was used for training acceleration. 
Adam was chosen for optimization with $\beta_{1} = 0.5$ and $\beta_{2} = 0.999$. Each mini-batch includes 4 samples in training phases. The learning rate was initialized as $0.000001$ and dropped by 0.1 every 40 epochs.

\begin{figure}[!tp]
\centering
\includegraphics[width=80mm]{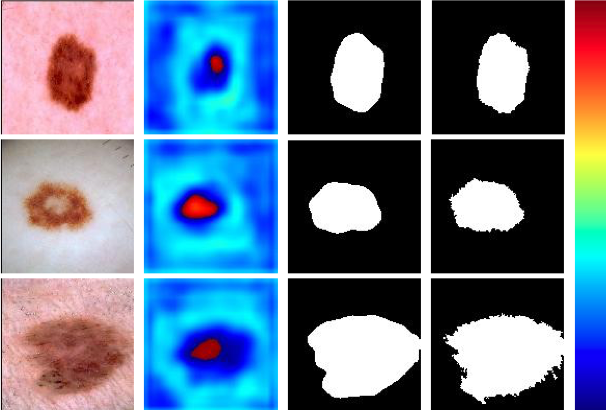}
\caption{Each row includes the original image, dilated rate map, predictions and ground truth from left to right. Note that red in heat map denotes a larger receptive field.}
\label{result1}
\vspace{-0.2cm}
\end{figure}

We first analyzed the influence of KAM, and showed the learned dilated rate maps at $32\times 32$ scale as in Fig.\ref{result1}. From the heat map, it is clear that the receptive field is expanded at the hole region, and the hole is disappeared in the final predictions, indicating the proposed KAM is able to solve the hole problem effectively. The receptive field around the boundary is slightly enlarged, which makes the prediction expand to fit the ground truth.
Therefore, context information of holes and boundaries is provided to help predict the class of local pixels, which alleviates hole as well as shrink problem and leads to better segmentation performance.

\begin{figure}[!tp]
\centering
\includegraphics[width=82mm]{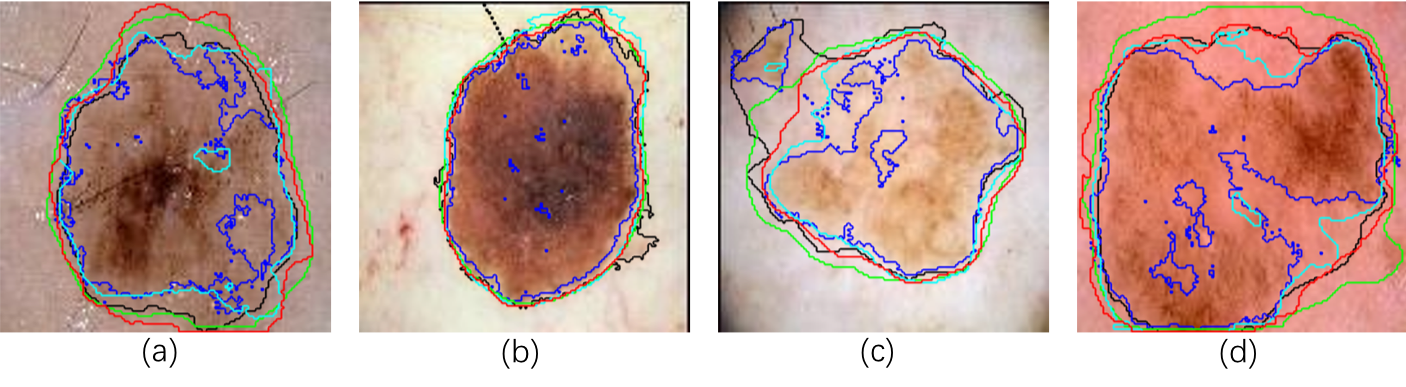}
\vspace{-0.2cm}
\caption{Examples of complementary network results in comparison with other methods. The ground truth is denoted in black. Results of \cite{ronneberger2015u}, \cite{sarker2018slsdeep}, \cite{yuan2017improving} and ours are denoted in blue, cyan, green, and red, respectively.}
\label{result2}
\vspace{-0.4cm}
\end{figure}
Then we assessed the qualitative performance of our complementary network by comparing it with state-of-the-art methods \cite{ronneberger2015u,sarker2018slsdeep,yuan2017improving}. We implemented these methods on our dataset and visualized four examples in Fig. \ref{result2}.
It is clear that our complementary network can appropriately fill holes, while holes existed in the results of methods \cite{ronneberger2015u,yuan2017improving}. Moreover, segmentation results obtained by the complementary network are better than that of other methods with smallest distance to the ground truth.

We further analyzed the performance of our complementary network by comparing it with methods \cite{ronneberger2015u,sarker2018slsdeep,yuan2017improving} under the setting of fully supervised and semi-supervised learning, respectively. Specifically, $3^{rd}$ to $6^{th}$ rows show fully supervised learning results with 1945 labeled images, while $8^{th}$ to $11^{th}$ rows report results with 649 labeled and 1296 unlabeled images for our method and results with 649 labeled images for other methods in Table \ref{pcam_table}. Under fully supervised setting, the proposed method shows superior performance with an improvement of 2.7\%, 0.2\%, 1.6\% in AC, 7.0\%, 0.5\%, 3.5\% in DI and 9.3\%, 1.0\%, 4.7\% in JA compared with the existing deep learning based methods \cite{ronneberger2015u,sarker2018slsdeep,yuan2017improving}, respectively. This result validated the proposed complementary network possesses superior ability to alleviate hole and shrink problems for skin lesion segmentation. 
Trained with the same labeled data, our semi-supervised method exhibited a significant increment in evaluation criteria compared with \cite{ronneberger2015u,sarker2018slsdeep,yuan2017improving}. The increment is more distinct than that of fully supervised methods, indicating our method can leverage unlabeled images effectively.


\begin{table}[tb]
\setlength{\abovecaptionskip}{0pt}
\setlength{\belowcaptionskip}{2pt}
\centering
\caption{Comparison results for menaloma segmentation.}
\label{pcam_table}
\scalebox{0.85}{\begin{tabular}{ c  c  c  c  c }
\toprule[1pt]
\multicolumn{5}{c}{\textbf{Fully supervised}}\\
\hline
Methods&AC (\%)&DI (\%)&JA (\%)&SE (\%)\\
Unet \cite{ronneberger2015u}& 92.3$\pm$0.3 & 79.4$\pm$0.4 & 68.3$\pm$0.6 & 83.6$\pm$0.6 \\
Sarker et al.\cite{sarker2018slsdeep}&94.8$\pm$0.1 & 85.9$\pm$0.3 & 76.6$\pm$0.6 & \textbf{87.9$\pm$0.5} \\
Yuan et al.\cite{yuan2017improving}& 93.4$\pm$0.2 & 82.9$\pm$0.8 & 72.9$\pm$0.9 & 85.6$\pm$0.6 \\
Our Method& \textbf{95.0$\pm$0.6} & \textbf{86.4$\pm$1.3} & \textbf{77.6$\pm$1.9} & 86.9$\pm$1.0  \\
\bottomrule[1pt]
\multicolumn{5}{c}{\textbf{Semi-supervised}} \\
\hline
Unet \cite{ronneberger2015u}& 91.0$\pm$0.2 & 75.9$\pm$0.3 & 63.8$\pm$0.3 & 82.3$\pm$0.2\\
Sarker et al.\cite{sarker2018slsdeep}&93.5$\pm$0.1 & 83.0$\pm$0.4 & 74.2$\pm$0.5 & \textbf{87.7$\pm$0.5}  \\
Yuan et al.\cite{yuan2017improving}& 91.4$\pm$0.3 & 77.8$\pm$0.9 & 66.4$\pm$1.1 & 84.7$\pm$0.7  \\
Our Method& \textbf{94.4$\pm$0.2} & \textbf{85.0$\pm$0.6} & \textbf{75.9$\pm$0.9} & 85.0$\pm$1.7  \\
\bottomrule[1pt]
\end{tabular}}
\end{table}

\section{Conclusion}
\vspace{-0.1cm}
In this paper, we propose a novel complementary network with adaptive atrous convolution (AAC), knowledge aggregation module (KAM) and mutual loss, for melanoma segmentation.
Our network is able to alleviate the hole and shrink problems existing in current methods. The proposed complementary network and mutual loss can be further extended to semi-supervised learning, which is significant for medical image segmentation due to the limited annotated data.
AAC and KAM can be flexibly transferred to other image segmentation tasks to adaptively enlarge receptive fields and boost the segmentation performance.



\end{document}